\def\year{2022}\relax
\documentclass[letterpaper]{article} % DO NOT CHANGE THIS
\usepackage{aaai22}  % DO NOT CHANGE THIS
\usepackage{times}  % DO NOT CHANGE THIS
\usepackage{helvet}  % DO NOT CHANGE THIS
\usepackage{courier}  % DO NOT CHANGE THIS
\usepackage[hyphens]{url}  % DO NOT CHANGE THIS
\usepackage{graphicx} % DO NOT CHANGE THIS
\usepackage{natbib}  % DO NOT CHANGE THIS AND DO NOT ADD ANY OPTIONS TO IT
\usepackage{caption} % DO NOT CHANGE THIS AND DO NOT ADD ANY OPTIONS TO IT
\DeclareCaptionStyle{ruled}{labelfont=normalfont,labelsep=colon,strut=off} % DO NOT CHANGE THIS
\usepackage{algorithm}
\usepackage{algorithmic}

\usepackage{booktabs}
\usepackage{multirow}
\usepackage{amsmath}
\usepackage{graphicx} 
\usepackage{amsfonts}
\usepackage{subfigure}
\usepackage{newfloat}
\usepackage{listings}
\floatstyle{ruled}
\newfloat{listing}{tb}{lst}{}
\floatname{listing}{Listing}
\title{SGEITL: Scene Graph Enhanced Image-Text Learning for Visual Commonsense Reasoning}
\makeatletter
\newcommand{\printfnsymbol}[1]{%
  \textsuperscript{\@fnsymbol{#1}}%
}

\author {
    Zhecan Wang \textsuperscript{\rm 1}\thanks{Equal Contribution},
    Haoxuan You \textsuperscript{\rm 1}\printfnsymbol{1},
    Liunian Harold Li \textsuperscript{\rm 2},
    Alireza Zareian \textsuperscript{\rm 1},
    Suji Park \textsuperscript{\rm 1},
    Yiqing Liang \textsuperscript{\rm 1},
    Kai-Wei Chang \textsuperscript{\rm 2},
    Shih-Fu Chang \textsuperscript{\rm 1}
}
\affiliations {
    \textsuperscript{\rm 1} Columbia University\\
    \textsuperscript{\rm 2} University of California, Los Angeles\\
    \textit{\{olinzhecanwang, haoxuanyou\}@gmail.com}
}
\usepackage{bibentry}
\begin{document}

\maketitle

\begin{abstract}
  Answering complex questions about images is an ambitious goal for machine intelligence, which requires a joint understanding of images, text, and commonsense knowledge, as well as a strong reasoning ability. Recently, multimodal Transformers have made a great progress in the task of Visual Commonsense Reasoning (VCR), by jointly understanding visual objects and text tokens through layers of cross-modality attention. However, these approaches do not utilize the rich structure of the scene and the interactions between objects which are essential in answering complex commonsense questions. We propose a
  \textbf{S}cene \textbf{G}raph \textbf{E}nhanced  \textbf{I}mage-\textbf{T}ext  \textbf{L}earning  ({\bf SGEITL}) framework to incorporate visual scene graph in commonsense reasoning. In order to exploit the scene graph structure, at the model structure level, we propose a multihop graph transformer for regularizing attention interaction among hops. As for pre-training, a scene-graph-aware pre-training method is proposed to leverage structure knowledge extracted in visual scene graph. Moreover, we introduce a method to train and generate domain relevant visual scene graph using textual annotations in a weakly-supervised manner. Extensive experiments on VCR and other tasks show significant performance boost compared with the state-of-the-art methods, and prove the efficacy of each proposed component.
\end{abstract}
\vspace{-5mm}
\section{Introduction}

\begin{figure}[t]
% \vspace{-10mm}
\begin{center}
\scriptsize
\scalebox{1}{
  \includegraphics[width=1\linewidth]{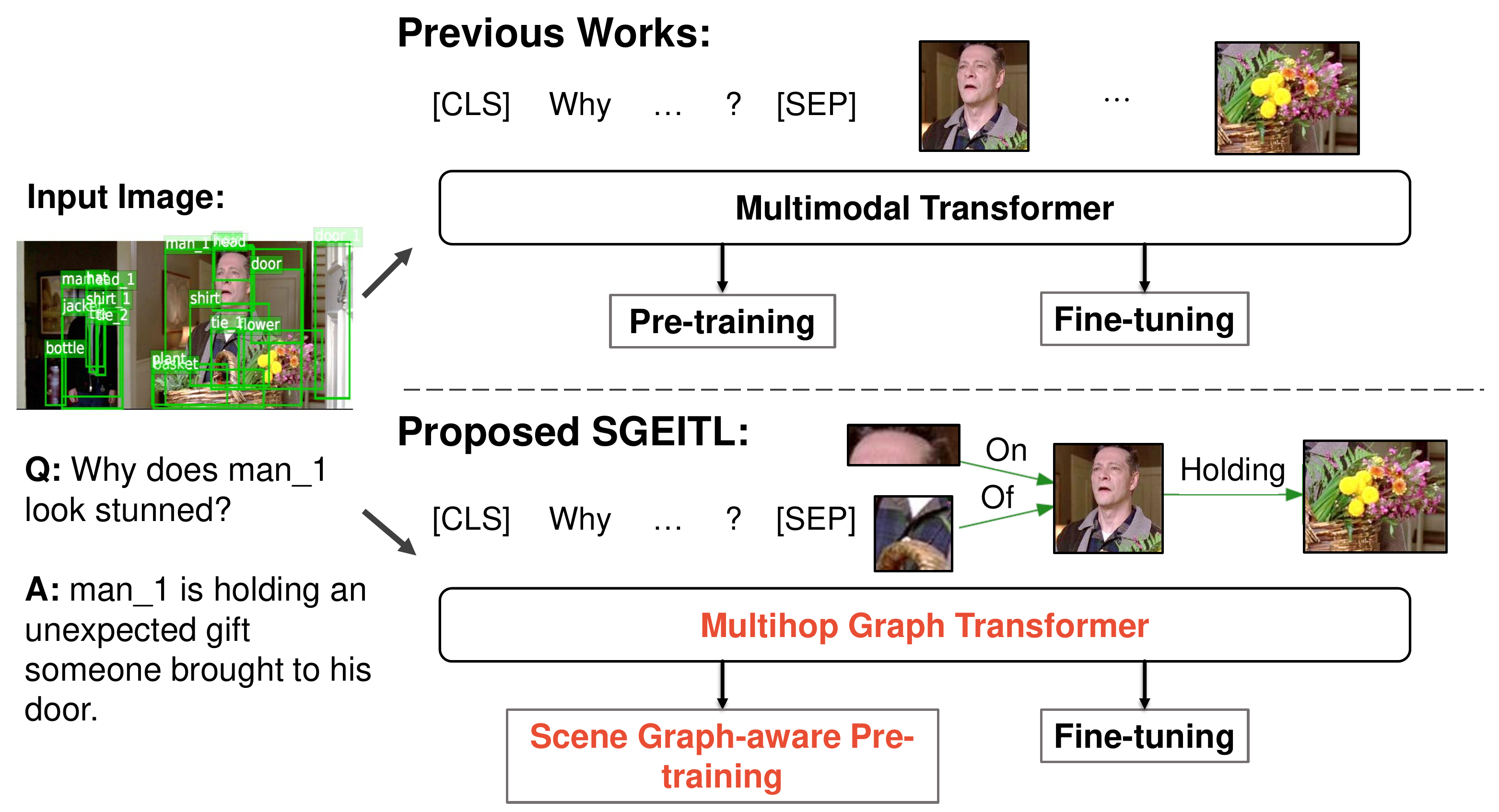}
}
\end{center}
% \vspace{-10mm}%Put here to reduce too much white space after your table
  %\caption{Overall Structure. Using scene graph into Vision-Language Transformer with graphical structure for Visual Commonsense Reasoning task}
  \vspace{-2mm}
  \caption{Overview. We propose to incorporate visual scene graph in VL model's pre-training, fine-tuning and conventional Transformer's structure for joint multimodal representations.}
\label{fig:overview}
\vspace{-6mm}
\end{figure}

Visual Commonsense Reasoning~\cite{zellers2019vcr} is a new addition to Vision-and-Language (VL) research, which has drawn significant attention in the past few years. Different from the conventional Visual Question Answering (VQA) task~\cite{goyal2017making}, VCR requires deeper understanding of the scene and commonsense knowledge. It also requires reasoning ability such as cause-effect and next-step prediction of the presented activity. The state-of-the-art (SOTA) performance on VCR has been improved by a series of recent works on Transformer-based VL models \cite{Su2020VL-BERT:,chen2020uniter,li2019visualbert}. In those models, visual object features from image and word embeddings from the image-question-answer pairs are jointly fed into the conventional Transformer model, which consists of several layers of multi-head attention both within each modality and across the two. 

In spite of their great performance, most existing Transformer-based models reduce the image into a bag of object features extracted using a pre-trained object detector~\cite{tan-bansal-2019-lxmert, chen2020uniter, Su2020VL-BERT:}. However, since the VCR task requires a comprehensive understanding of the visual scene and the commonsense reasoning beyond that, object information alone may not be sufficient to understand visual scenes. Therefore, a more comprehensive visual representation along with the paired text prompt is essential for progress in this field. 

A scene graph represents an image as objects and their interactions, providing a structured understanding of the visual scene. 
Due to its compact, yet comprehensive representation, it has been used for several applications, such as image retrieval~\cite{Johnson_2015_CVPR}, image captioning~\cite{Yang_2019_CVPR}, image synthesis~\cite{johnson2018image}, and visual question answering~\cite{VQA}. Nevertheless, key limitations have prevented us from utilizing scene graphs for VCR. Firstly, it is unclear how the top-performed Transformer-based methods that assume a set of visual and textual tokens as input can handle the information present in a graph structure. Secondly, existing pre-training tasks of VL models mainly use naive random masking when calculating pre-training loss. Such masking steps \cite{Su2020VL-BERT:, li2020oscar} lack the ability to consider graph connections if tokens are from visual scene graphs. Thirdly, the only public dataset with sufficient annotation to train scene graph generation (SG) models is Visual Genome (VG)~\cite{krishna2017visual} which is heavily biased in relationship classes \cite{zellers2018scenegraphs} and its classes also suffer from a severe semantic gap compared to the questions and answers in VCR.

In this work, we aim to address the three aforementioned challenges by incorporating visual scene graph into Transformer's model structure, pre-training and fine-tuning. Those adjustments in VL models' training pipelines are model-agnostic, and could be unified to a Scene Graph Enhanced Image-Text Learning (SGEITL) schema to assist popular VL models' pre-training and fine-tuning, as shown in Fig. \ref{fig:overview}. 

% in a single framework, {\bf G}raph-based {\bf I}mage-{\bf T}ext {\bf R}epresentations {\bf L}earning ({\bf GITRL}), as shown in Fig. \ref{fig:overview}. 

SGEITL takes text tokens and visual scene graph tokens as input. Then the proposed multihop graph structure is applied to existing Transformer's attention to learn joint multimodal representations through a set of pre-training tasks. Modified on top of current VL Transformers, multihop graph Transformer can dynamically adjust the attention value between tokens within multiple hops based on mutual distance in a scene graph. In addition, for pre-training,  we propose a scene-graph-aware pre-training method with triplet-based masking. Under this strategy, each triplet (subject/object/predicate) of visual scene graphs can have at most one component masked. Compared with random masking strategy in previous works, our method preserves the necessary local context for the masked node in each triplet unit and also emphasizes the semantic and structural difference among nodes in the scene graph with three prediction heads for subject, object and predicate predictions respectively. Lastly, for solving the problem of limited scene graph annotation and existing biased distribution of scene graph dataset, we further propose an innovative SGG method, independent from VL Transformer models, driven by the weak supervision of parsed text of each image-text pair.

The main contributions of our paper are twofolds. 
\vspace{-1mm}
 \renewcommand{\labelenumii}{\Roman{enumii}}
 
  \begin{enumerate}
\item To the best of our knowledge, this is the first work to demonstrate that the structure information in visual scene graph is helpful for complex semantic visual question answering task, such as the VCR task. 
\vspace{-1mm}
\item We systematically introduce multiple ways of enhancing current VL models with visual scene graphs, which enable a more structural understanding towards visual scene and hence facilitate multimodal learning. Our schema includes the following components and is generalizable to different VL models.
% \item We introduce GITRL, a novel framework that utilizes scene graph in the visual question answering task. Our framework includes the following components and is easy to generalize to other dataset.
  \vspace{-1mm}
  \begin{enumerate}
  \item Our multihop graph structure inherits from the conventional Transformer could explicitly model the multihop reasoning on visual scene graph.
  \vspace{-1mm}
  \item  Our scene-graph-aware pre-training method takes into consideration of visual scene graph's structure and could assist VL models for learning better visual representation.
  \vspace{-1mm}
  \item  Our new SGG method, Text-VSPNet can learn to generate scene graphs even for datasets without scene graph annotation. With weakly-supervision from text, the model can generate semantically-rich and target domain-relevant visual scene graphs.
\end{enumerate}

\end{enumerate}

\section{Related Work}

\begin{figure*}[htpb]
\vspace{-4mm}
\begin{center}
\scriptsize
\scalebox{0.8}{
  \includegraphics[width=0.90\linewidth]{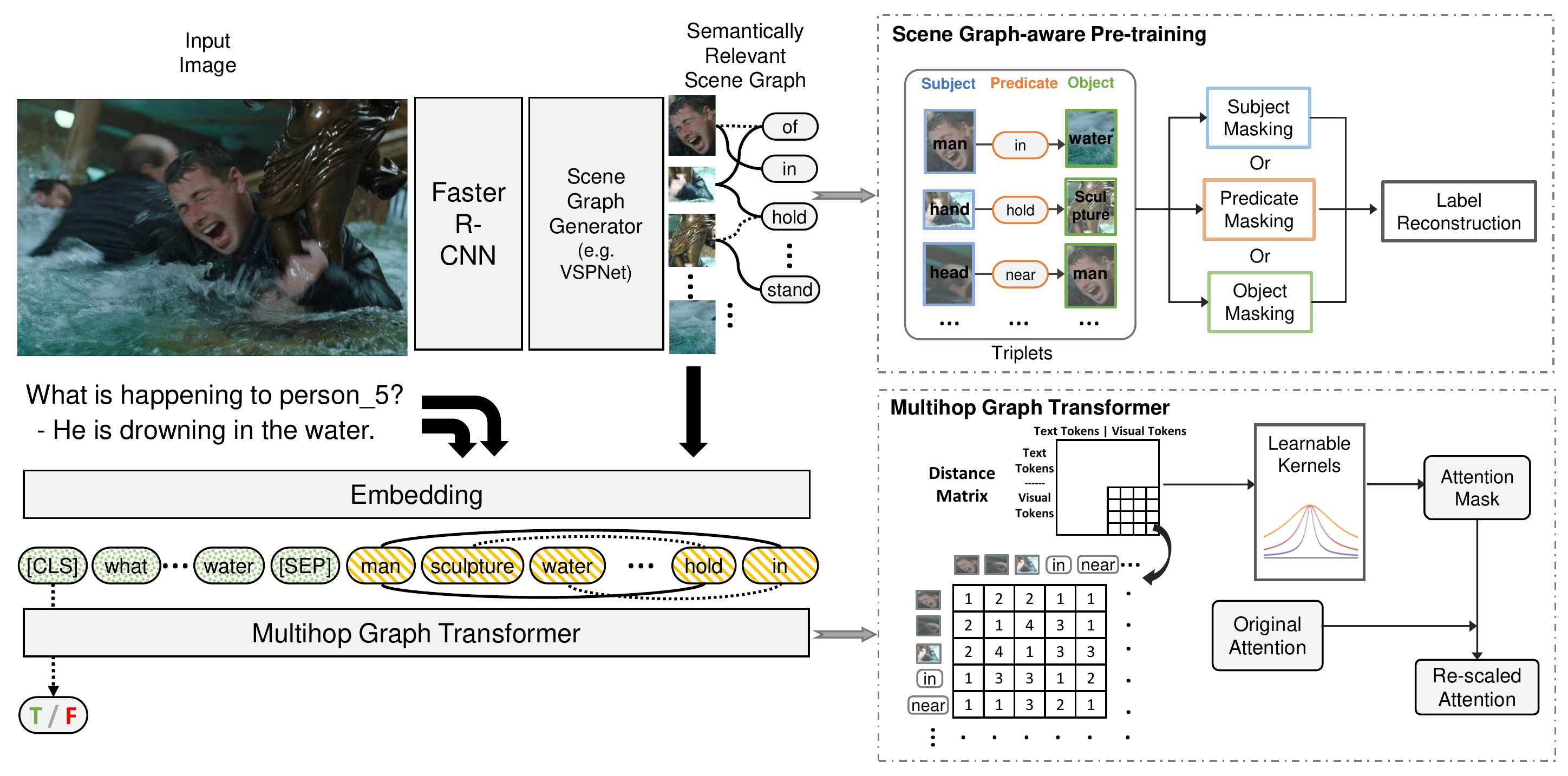}
}
\end{center}
\vspace{-4mm}%Put here to reduce too much white space after your table
  %\caption{Overall Structure. Using scene graph into Vision-Language Transformer with graphical structure for Visual Commonsense Reasoning task}
  \caption{Pipeline. We first extract a scene graph from a given image, using either a conventional SGG method, or the proposed Text-VSPNet. Then we employ the multihop graph Transformer to tackle the scene graph input with regularized attention mechanism among multiple hops. For pre-training, a novel scene-graph-aware pre-training schema is utilized to conduct masking on triplet-level.}
\label{fig:pipeline}
\vspace{-6mm}%Put here to reduce too much white space after your table
\end{figure*}

\subsection{Vision and Language Representations}

\paragraph{Multimodal Transformers:} Combining vision and language is essential for various tasks such as visual question answering \cite{VQA,balanced_binary_vqa,balanced_vqa_v2} and visual reasoning \cite{suhr-etal-2019-corpus,zellers2019vcr}. The emerging trend of multimodal Transformers has shown promising progress in these fields, where general-purpose models are pre-trained on image-caption pairs, and then fine-tuned on downstream by transferring their rich representations from pre-training. These models are simpler, yet more effective and versatile~\cite{lu2019vilbert,tan-bansal-2019-lxmert,li2019visualbert,Su2020VL-BERT:,chen2020uniter,li2020weakly}. The idea behind is to extend the language Transformer model (\textit{e.g.} BERT~\cite{devlin-etal-2019-bert}) to vision by adding visual tokens. The model in most cases is exactly the same as BERT. Some variants such as LXMERT~\cite{tan-bansal-2019-lxmert} employ a two-stream architecture which applies Transformers separately on each modality, followed by a multimedia Transformer across modalities. In contrast to these approaches, our proposed multihop graph Transformer takes visual scene graphs as the visual input instead of a bag of object features, and incorporates the scene structure through a graph-based attention mechanism.
\paragraph{Multimodal Pre-training:}
 Since \cite{devlin-etal-2019-bert, Radford2018ImprovingLU}, pre-training on general representations followed by transferring knowledge on downstream tasks has been a very popular strategy for many different tasks. Besides text-only pre-training, former works \textit{e.g.} \cite{li2019visualbert, tan-bansal-2019-lxmert, lu2019vilbert} use a similar set of multimodal pre-training methods and demonstrate their efficacy in downstream multimodal tasks \cite{VQA,balanced_binary_vqa,balanced_vqa_v2, suhr-etal-2019-corpus,zellers2019vcr}. Most of them apply naive random masking and expect to learn strong multimodal representations. Following them, later methods like \cite{li2020oscar, chen2020uniter} focus and improve on semantic alignments between vision and text domains via variant pre-training methods related to using  contrastive loss, word region alignment and object tags as anchors. However, these explicit alignments are limited to syntactic matching of tokens without considering structure information. A recent method \cite{yu2020ernie-vil} takes a step further to obtain scene-related graphs parsed from the paired texts during pre-training. However, the texts are mostly short, having very limited visual descriptive tokens related to the scene. This would surely bring an insufficient description of the visual scene by the solely text-parsed graph. Also, it still inherits from the former to use bags of tokens for the visual representation. This would cause inconsistency in representations between vision and language, hinder the learning of multimodal alignment and cannot ensure to have sufficient local context during reconstruction. Differently, in this work, we focus on incorporating visual scene graphs from images during pre-training to help models adapt to the structural scene domain.

\paragraph{Scene Graph Generation:} \textbf{S}cene \textbf{G}raph \textbf{G}eneration (\textbf{SGG}) has attracted much attention since proposed in \cite{xu2017scenegraph} and shown potential benefits for several downstream visual reasoning tasks \cite{krishna2018referring, johnson2018image,jiang2019dualvd, shi2019explainable, Yang_2019_CVPR,zhu2020mucko}. The goal of SGG is to take an image and extract a set of objects and pairwise interactions, which will form a graph where predicates are either edges~\cite{yang2018graph} or nodes~\cite{Zareian_2020_CVPR}. Although most SGG methods need intensive supervision to be trained, recently VSPNet~\cite{Zareian_2020_CVPR} is proposed as a more generalized form of SGG, which does not require bounding box supervision. Although VSPNet was originally trained on Visual Genome (VG), considering its capability of learning from weak supervision, we adopt VSPNet to the task of VCR to generate semantically rich and task-related scene graphs by pre-training on Visual Genome~\cite{krishna2017visual} and then fine-tuning on the text annotations on VCR~\cite{zellers2019vcr}. 
\vspace{-1mm}
\section{Scene Graph Enhanced Image-Text Learning}
\vspace{-0.3mm}
In this section, we will first explain the high-level architecture of how we equip VL models with visual scene graphs in Sec. \ref{sec3.1}. Then we introduce the multihop graph Transformer in Sec. \ref{sec3.3}, scene-graph-aware pre-training strategy in Sec. \ref{sec3.4}, and explain how to generate visual scene graph semantically-relevant to text domain in Sec. \ref{sec3.5}.

% In this section, we will first explain the high-level architecture of the proposed GITRL framework in Sec. \ref{sec3.1}. Then we introduce the multihop graph Transformer in Sec. \ref{sec3.3}, graph-based multimodal pre-training strategy in Sec. \ref{sec3.4}, and explain how to generate visual scene graph semantically relevant to text domain in Sec. \ref{sec3.5}.

\subsection{Model Overview}
\label{sec3.1}

The overall pipeline of our approach is illustrated in Fig. \ref{fig:pipeline}.For input, we take generated visual scene graph from an image and text tokens as input, and extract joint multimodal representations. For preserving the graph connections between visual nodes, we upgrade the conventional Transformer with the proposed multihop graph for learning multihop feature aggregation. To accommodate the visual scene graph's structure in input, we introduce a novel triplet-based masking strategy in the pre-training phase. Moreover, we introduce a new model-agnostic way to obtain visual scene graphs that are more relevant to the VCR domain by training an SGG model directly on VCR text annotations.

% The overall pipeline of our approach is illustrated in Fig. \ref{fig:pipeline}. We propose the {\bf G}raph-based {\bf I}mage-{\bf T}ext {\bf R}epresentations {\bf L}earning ({\bf GITRL}) framework, which takes generated visual scene graph from an image and text tokens as input, and extracts joint multimodal representations. For preserving the graph connections between visual nodes, we propose the multihop graph Transformer for learning multihop feature aggregation. To accommodate the visual scene graph input to the text domain, we introduce a novel unified triplet-based masking strategy in the pre-training phase. Moreover, we introduce a new way to obtain visual scene graphs that are more relevant to the VCR domain by training an SGG model directly on VCR text annotations.

The key difference between scene graph enhanced VL learning and previous multimodal VL learning is twofold. (1) Each image is modeled as a scene graph, consisting of object features, relation features and connectivity information. (2) With bringing visual scene graph into previous VL learning frameworks and our proposed adjustments in three perspectives, we not only transform the generated graph to be domain-relevant via weak supervision but also effectively exploit the relation features and graph structures through scene-graph-aware pre-training and modeling.

\vspace{-3mm}
\subsection{Scene Graph as Visual Input}
\label{sec3.2}
Each input sample consists of a text segment and a scene graph represented as a set of objects and relations and their connections. For the text embedding layer, following BERT \cite{devlin-etal-2019-bert}, we tokenize the input sentence into a sequence of WordPieces \cite{wu2016google} and obtain its BERT Embedding. For the scene graph embedding layer, we represent each object and relation by two types of features: (1) position features, (2) visual features. Position features of objects and relations are respectively transformed from bounding box coordinates and union box coordinates. For visual features of objects, we take the region of interest (ROI) features from the object detector; for visual features of relations, we take the relation features before inputting into the final prediction layer in SGG models. Besides, since ROI features and relation features are in different semantic space, two fully-connected layers are applied individually to object visual features and relation visual features.

The obtained text embeddings and visual embeddings are fed into the Transformer-based model.

\subsection{Multihop Graph Transformer}
\label{sec3.3}

In previous VL Transformer models, since the visual input is either a sequence of objects~\cite{lu2019vilbert, li2019visualbert} or pixels~\cite{huang2020pixel}, they directly use the conventional Transformer \cite{vaswani2017attention}, in which every token (both visual and text) can freely attend to each others' belittling local connections between relevant objects. However, when the input includes a scene graph, the attention should favor the local interactions between connected nodes (objects and predicates) in graph structure since they are essential to the entire visual scene understanding. Toward such motivation, we modified the conventional Transformer with multihop graph attention mechanisms. So the improved Transformer can dynamically adjust the attention weight between visual tokens within multiple hops.  

We denote a generated visual scene graph as $G(V, E)$, where $V$ denotes vertices of objects and predicates and $E$ denotes edges connecting them. Following \cite{yang2018graph}, besides edges in $G(V, E)$, we also add skip edges between all objects to allow direct information flow among objects and get an enhanced scene graph $G^\prime(V, E^\prime)$. Given a sequence of input tokens $\{x_i\}^{i=n}_{i=1}$ (including text tokens, objects tokens and predicates tokens), we first pre-compute a distance matrix $D$ based on $G^\prime(V, E^\prime)$. The distance between $i$-th token and $j$-th token is defined as the number of edges (hops) in the shortest path. It's noted that the distance between visual tokens and text tokens is always set to 1 because we want the two modalities to be fully connected. The distance between text tokens is also set to 1 to prevent from disrupting the knowledge learned during BERT pre-training.  
\vspace{-0.7mm}
\begin{equation*}
     D_{ij} = \begin{cases}
Distance(i,j)& \text{$x_i$ and $x_j$ in $G^\prime(V, E^\prime)$,} \\
1& \text{$x_i$ and $x_j$ in different modalities,}\\
1& \text{$x_i$ and $x_j$ in text modality.}\\
\end{cases}
\end{equation*}

Following the input sequence, a $L$-layer multihop graph attention mechanism is applied. Inside each layer, multiple attention heads are included. At each attention head, the output of previous layer is treated as input to Key $W^K$ and Query $W^Q$ and Value $W^V$ to be projected to hidden dimension. 

\vspace{-4mm}
\begin{equation*}
     Q=H_{in}W^Q, K=H_{in}W^K, V=H_{in}W^V.
\end{equation*}

Then the attention matrix is computed by a scaled dot-product between $Q$ and $K$. Similar to  \cite{Zareian2020LearningVC, ahmad2020gate}, we applied a binary mask to zero out the attention values beyond $h$ hops.  

\vspace{-2mm}
\begin{equation}
     A=softmax\left(\frac{QK^{T}}{\sqrt{d_k}}+M\right),
\end{equation} 
\begin{equation}
    \mbox{where }  M_{ij}=\begin{cases}
0& D_{ij}\leq h, \\
-\infty& D_{ij}> h.
\end{cases}
\end{equation}

After getting the attention matrix $A$, the next step of the conventional Transformer is to multiply it with value $V$ to aggregate the features as output. However, in visual scene graph, it's broadly proved that the predicates largely depend on the objects they link to \cite{zellers2018scenegraphs, chen2019knowledge}. Based on above finding, we further hypothesize that the attention between closer nodes in scene graph should be emphasized and vice versa. That also coheres with the conclusion drawn by \cite{ahmad2020gate} when they utilize dependency graph in text Transformer to help cross-lingual relation and event extraction tasks. Therefore, in our multihop graph attention, built on top of the conventional attention mechanism, a monotonically decreasing function $F:D \rightarrow\mathbb{R}^1$ is introduced to generate an additional attention mask for re-scaling the original attention values in $A$. And one more normalization function $\sigma$ is cascaded to make sure the sum to be 1. 

\vspace{-4.5mm}
\begin{equation} \label{eq:3}
     H_{out}=\sigma(F(D)\cdot A) V,
\end{equation} 
\begin{equation} \label{eq:4}
\sigma(e_{ij})=\frac{e_{ij}}{\sum_{j=1}^{n}e_{ij}}.
\end{equation} 

Considering that different heads should have different functionalities, instead of a fixed handcrafted function as in \cite{ahmad2020gate}, we propose to use individual parametric kernel in each head with learnable parameters. Based on experiments, we choose the rational quadratic (RQ) kernel, which is similar to Gaussian kernel but with a smoother prior. 
% The lengthscale $l$ and the scale-mixture $\alpha$ are learnable to control the smoothness/sharpness of the function.

Furthermore, one important property of any scene graph is that it's always a bipartite graph where the edges are only between objects and predicates. Even though we add the skip connections between objects for information flow, the neighbor distributions of objects and predicates still vary a lot. For example, within scene graph, the second-hop neighbors of predicates and objects are always predicates and objects respectively. With the motivation to disentangle them, we employ two learnable kernels separately for objects and predicates. 
\vspace{-3mm}
\begin{equation}
     F(D)_{ij}=\begin{cases}
     (1+\dfrac{(D_{ij}-1)^2}{2\cdot \alpha_o \cdot l^2_o})^{-\alpha_o}& \text{$x_i$ $\in$ objects}\\
     (1+\dfrac{(D_{ij}-1)^2}{2\cdot \alpha_p \cdot l^2_p})^{-\alpha_p}& \text{$x_i$ $\in$ predicates}\\
     1 & \text{else}
     \end{cases}
\end{equation} 
where $\alpha_o$ ($\alpha_p$) and $l_o$ ($l_p$) are $1\times1$ learnable scale-mixture parameter and $1\times1$ length-scale parameter for objects (predicates). It's noted that in order to upgrade the conventional attention to multihop graph attention, we only need to add four parameters per attention head which is quite efficient.

\begin{figure}[t]
% \vspace{-4mm}
\begin{center}
\scriptsize
\scalebox{1}{
  \includegraphics[width=0.95\linewidth]{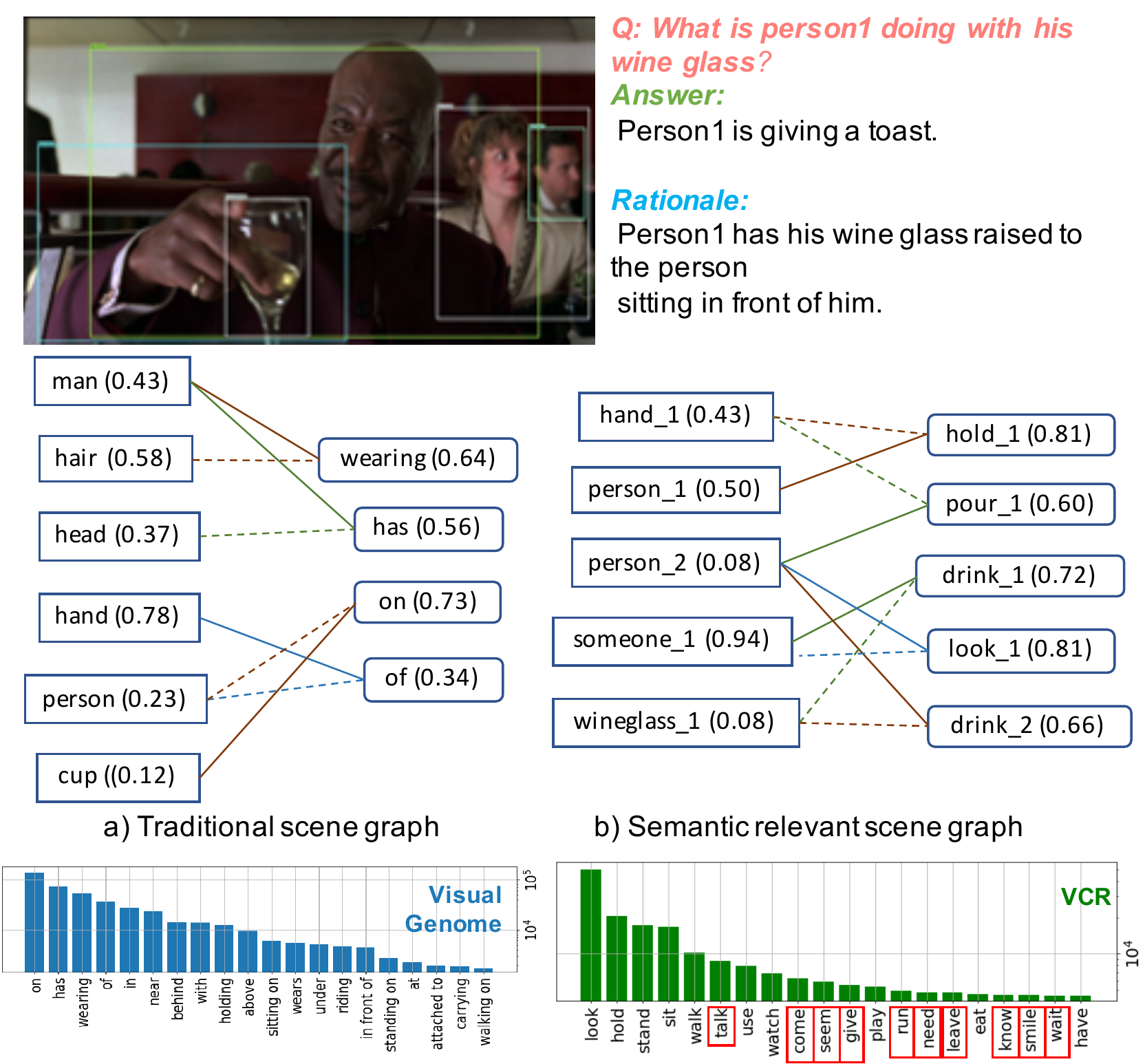}
}
\end{center}
% \vspace{-7mm}%Put here to reduce too much white space after your table
  \caption{Comparison between traditional and semantically-relevant scene graphs (Best viewed by zoom-in).}
\label{fig:4}
\vspace{-5mm}
% \vspace{-7mm}%Put here to reduce too much white space after your table
\end{figure}

\subsection{Scene-Graph-Aware Pre-training}
\label{sec3.4}

Pre-training has become a conventional practice for multimodal frameworks to fuse different modalities. Former VL models \cite{chen2020uniter, li2020oscar, li2019visualbert, gan2020large} have proved that large-scale pre-training on image-text data would significantly benefit the downstream tasks. However, most of the current pre-training methods rely on naive random masking ignoring the semantic difference of tokens, structure information and local context. For instance, in a triplet, $ [man, on, floor]  (subject, relationship, object) $, with random masking, it is possible that both the subject and the object are masked out. Under this situation, it is challenging to find sufficient local context to reconstruct the missing token. Without constraint of context, relationship tokens like $ [on] $ would have too many possible combinations of subject and object. To solve this problem, we incorporate visual scene graphs in pre-training steps to preserve the structural context during masking. Visual scene graphs are consisted by triplets. Each triplet has two entities \textit{i.e.} subject as well as object and also one relationship connecting those two. In scene-graph-aware pre-training, we use visual scene graphs as the guidance to selectively mask tokens conditioning on triplet to ensure that sufficient local context would be preserved for reconstructing the masked nodes. This facilitates the model to learn the correlation in neighbor.

The recent work \cite{yu2020ernie-vil} claims to obtain "scene graph" parsed from the text and utilize it during pre-training. However, its parsed graph should be more accurately regarded as scene-related graph since it does not closely reflect the object interaction and spatial relationship in the image. It is also not consisted of the conventional triplet $(subject, relationship, object)$ thus it reflects more of loose text dependency instead of close visual object interactions. Differently, our visual scene graph is directly extracted from the image. It can closely represent the interaction between objects and spatial relationships. Furthermore, following the triplet partition, our masking strategy is relationship-centric focusing on interactions between objects \cite{shi2019simple}. For each image-text pair, \cite{yu2020ernie-vil}'s graph information is never inputted into the VL model and only used for masking text tokens for calculating pre-training loss. Due to the loose text dependency connection of its graph, the pre-training would not allow model to better learn the visual structure of the scene. However, differently, we do feed the visual scene graph into the VL model and even directly mask to modify the visual object input. We have three prediction tasks corresponding to three masked semantic roles (subject, object and relationship). Each visual sample would be equally possible to be assigned for only one of the three types of prediction tasks. After assigned, for each sample, all the visual objects of the semantic role corresponding to the assigned prediction task, we would randomly select $30 \%$ of them and assign a special \textit{[MASK]} token \textit. Therefore, for any triplet, only one of the three nodes would be masked leaving sufficient context information for modeling. Furthermore, we also have three different prediction heads corresponding to the three prediction tasks. This explicit separation allows the model to learn the semantic difference among nodes and better understand the structural knowledge. Under this mechanism, we introduce the pre-training loss of {\bf M}asked {\bf N}ode {\bf M}odeling ({\bf MNM}).\vspace{-2mm}
\begin{equation}
\mathcal{L}_{\mathrm{MNM}}(\theta)= \mathcal{L}_{s b j}(\theta) + \mathcal{L}_{o b j}(\theta) + \mathcal{L}_{r e l}(\theta),
\end{equation}where $\theta$ is the trainable parameters. The model would be supervised to predict the masked node of each triplet, $\left\langle\mathbf{n}_{s_{i}}, \mathbf{n}_{o_{i}}, \mathbf{n}_{r_{i}}\right\rangle$ based on the observation of the other two unmasked nodes in the triplet, other surrounding
nodes $\mathbf{N}$ and all tokens from the other modality V, by minimizing the negative log-likelihood:\vspace{-2mm}\begin{equation*}
\mathcal{L}_{s b j}(\theta) = -\mathbb{E}_{(\mathbf{n}, \mathbf{v}) \sim D} \log P_{\theta}\left(\mathbf{n}_{\mathbf{s_{i}}} \mid \mathbf{n}_{\backslash \mathbf{s_{i}}},
\mathbf{n}_{\mathbf{o_{i}}},
\mathbf{n}_{\mathbf{r_{i}}},
\mathbf{N},
\mathbf{V}\right),
\end{equation*}\begin{equation*}
\mathcal{L}_{o b j}(\theta) = -\mathbb{E}_{(\mathbf{n}, \mathbf{v}) \sim D} \log P_{\theta}\left(\mathbf{n}_{\mathbf{o_{i}}} \mid \mathbf{n}_{\backslash \mathbf{o_{i}}},
\mathbf{n}_{\mathbf{s_{i}}},
\mathbf{n}_{\mathbf{r_{i}}},
\mathbf{N},
\mathbf{V}\right),
\end{equation*}\begin{equation*}
\mathcal{L}_{r e l}(\theta) = -\mathbb{E}_{(\mathbf{n}, \mathbf{v}) \sim D} \log P_{\theta}\left(\mathbf{n}_{\mathbf{r_{i}}} \mid \mathbf{n}_{\backslash \mathbf{r_{i}}},
\mathbf{n}_{\mathbf{s_{i}}},
\mathbf{n}_{\mathbf{o_{i}}},
\mathbf{N},
\mathbf{V}\right),
\end{equation*}
where $i$ is a sampled location for masking nodes, $s$ represents subject, $o$ represents object and $r$ represents their relationship.

% Similar to the visual input, we could further use the same semantic role labelling parser \cite{shi2019simple} to parse the text input into a scene-related graph. Different from \cite{yu2020ernie-vil}, it would also have the same triplet structure of the same three semantic roles as the visual scene graph. We can utilize the shared structure information to conduct more detailed semantic alignment between the two modalities. During multimodal pre-training, only one prediction task would be assigned for each image-text pair. Reconstruction of masked tokens of the same semantic role from two modalities together would facilitate the two graphs to better align and form a unified graph representing the scene. The final pre-training loss would be the sum of masked node modeling losses from both visual scene graph, $\mathcal{L}_{\mathrm{MNM_v}}(\theta)$ and text-parsed graph, $\mathcal{L}_{\mathrm{MNM_t}}(\theta)$:
% \vspace{-2mm}
% \begin{equation}
% \mathcal{L}_{\mathrm{Pre-training}}(\theta)= \mathcal{L}_{MNM_v}(\theta) + \mathcal{L}_{MNM_t}(\theta)
% \end{equation}

\subsection{Weakly Supervised Scene Graph Generation}
\label{sec3.5}
Besides modifying current VL models with scene graph related pre-training and structure modification, we also explore to improve visual scene graph at input level. SGG methods are typically trained on the Visual Genome (VG) dataset, which is the only large-scale source for SGG supervision. However, due to the limited and biased distribution of VG, the conventionally produced scene graphs sometimes may not be ideally semantically-relevant for VCR questions. In order to produce more useful scene graphs, inspired by \cite{Zareian_2020_CVPR}, we innovatively make the first attempt to train SGG with weak supervision from text data. 

% domain with diverse and semantic-heavy predicates such as the VCR task.
% from the target domain's text

VSPNet \cite{Zareian_2020_CVPR} is a graph-based neural network architecture that takes object proposal features as input and creates a bipartite graph consisting of entities (objects) and predicates (visual relations). The original VSPNet was originally tested on VG dataset to obtain its supervision signal via alignment against labeled ground truth graph. However, most image-text datasets do not provide scene graph annotation, generating scene graphs for those unlabeled datasets is a difficult problem but also very important. It is even more challenging for highly semantic dataset like VCR. Considering this, we make the first attempt to utilize the paired text prompts in image-text datasets to generate weak supervision signal to train our scene graph generator, Text-VSPNet. We use a semantic parser \cite{shi2019simple} to extract verbs, nouns, and semantic roles from the text prompts and then use a coreference resolution model \cite{lee2017end} to merge 
corresponding noun nodes of such graphs. This is followed by a set of rule-based post-processing steps to create clean graphs as the "ground truth" for training Text-VSPNet. With our flexible weakly-supervised training, the generated scene graph can obtain richer classes of objects and predicates. More importantly, when using the target domain's text prompts as the supervision signal, the generated scene graph is also more semantically-relevant to the target domain's text and images, as shown in the bottom right of Fig. \ref{fig:4}. This especially facilitates the utilization of scene graph on highly semantic dataset like VCR. In VCR, we use both questions and answers from the train dataset only to extract pseudo-ground-truth graphs. This process results in 100K images with ground truth semantic graphs that include 672 object classes and 521 predicate classes, which comprise the most frequent words in VCR QAs, and hence are much more likely to be relevant to VCR questions than VG annotations, shown in Fig. \ref{fig:4}. 

% We use this large, albeit noisy dataset to train the VSPNet model, and use the trained model to extract scene graphs for feeding into our SGEITL model.

In practice, we further notice that Text-VSPNet is prone to data frequency bias, and it does not learn infrequent classes well. To mitigate this problem, we augment the cross-entropy entity and predicate classification loss terms in VSPNet by focal loss \cite{lin2017focal} and class-balanced loss \cite{cui2019class}. This results in much more diverse predictions that we empirically found essential for VCR.
\vspace{-2mm}
\section{Experiments}

In this section, we analyze different components of our framework and compare the performance with the SOTA methods. Additionally, visualization is shown to illustrate the intuition behind our model.

% \begin{table*}[htpb]
% \begin{center}
% \scalebox{0.8}{
% \renewcommand\tabcolsep{5pt} % 调整表格列间的长度
% \begin{tabular}{ccccccc|ccc}
%     \toprule
%      &    &\multirow{2}{*}{Model} & \multirow{2}{*}{SceneGraph} &  
%      \multirow{2}{*}{SceneGraph+}  &
%      \multirow{2}{*}{Pretrain-V} & \multirow{2}{*}{HopTrans} &  \multicolumn{3}{c}{Accuracy} \\
%     %\cline{7-9} \cline{10-12}
%     & & & & & &   &Q$\rightarrow$A &QA$\rightarrow$R&Q$\rightarrow$AR \\
%     \midrule
%     %\cline{7-9} \cline{10-12}
%     1 &  & Baseline &   &  &  &  & 72.9 & 75.3 & 54.9 \\
%     \midrule
%     2 & & \multirow{5}{*}{SGEITL} & $\surd$ &  & &  &  73.1 & 75.6 & 55.3\\
%     % \midrule
%     % \cline{4-10} \cline{10-12}
%     % 3 & &  & $\surd$ &$\surd$  &  & & 73.8 & 76.3 & 56.3\\
%     % \midrule
%     3 & &  & $\surd$ &$\surd$  &  & & 73.5  & 76.2 & 56.0\\
%     % \midrule
%     4 & &   &$\surd$  &$\surd$  & $\surd$ & & 74.4  & 76.9 & 57.2\\
%     5 & &   &$\surd$  &$\surd$  &$\surd$ & $\surd$ & 74.8  & 77.2 & 57.7 \\
%     % \midrule

%     \bottomrule
% \end{tabular}
% }
% \end{center}
% \caption{Ablation results of SGEITL  on VCR validation set.  SceneGraph represents adding conventional predicates from SGG trained on VG dataset. Pretrain-V means  scene graph-aware pre-training on visual scence graph.  HopTrans means applying multihop graph on the input. SceneGraph+ means adding more domain-relevant predicates from SGG weakly supervised on annotated text input.)}
% \label{abalation_table}
% \vspace{-4mm}

% \end{table*}

\begin{table}[t]
\begin{center}
% \small
% \renewcommand\tabcolsep{2pt} % 调整表格列间的长度
\scalebox{0.6}{
\begin{tabular}{ccccccc|ccc}
    \toprule
     &    &\multirow{2}{*}{Model} & \multirow{2}{*}{SceneGraph} &  
     \multirow{2}{*}{SceneGraph+}  &
     \multirow{2}{*}{Pretrain-V} & \multirow{2}{*}{HopTrans} &  \multicolumn{3}{c}{Accuracy} \\
    %\cline{7-9} \cline{10-12}
    & & & & & &   &Q$\rightarrow$A &QA$\rightarrow$R&Q$\rightarrow$AR \\
    \midrule
    %\cline{7-9} \cline{10-12}
    1 &  & Baseline &   &  &  &  & 72.9 & 75.3 & 54.9 \\
    \midrule
    2 & & \multirow{5}{*}{SGEITL} & $\surd$ &  & &  &  73.1 & 75.6 & 55.3\\
    % \midrule
    % \cline{4-10} \cline{10-12}
    % 3 & &  & $\surd$ &$\surd$  &  & & 73.8 & 76.3 & 56.3\\
    % \midrule
    3 & &  & $\surd$ &$\surd$  &  & & 73.5  & 76.2 & 56.0\\
    % \midrule
    4 & &   &$\surd$  &$\surd$  & $\surd$ & & 74.4  & 76.9 & 57.2\\
    5 & &   &$\surd$  &$\surd$  &$\surd$ & $\surd$ & 74.9  & 77.2 & 57.8 \\
    % \midrule

    \bottomrule
\end{tabular}
}
\end{center}
\caption{Ablation results of SGEITL  on VCR validation set.  SceneGraph represents adding conventional predicates from SGG trained on VG dataset. Pretrain-V means  scene-graph-aware pre-training on visual scence graph.  HopTrans means applying multihop graph on the input. SceneGraph+ means adding more domain-relevant predicates from SGG weakly supervised on annotated text input.}
\label{abalation_table}
\vspace{-4mm}
\end{table}

\vspace{-2mm}
\subsection{Dataset}
\label{4.1}
Details of dataset is attached in the supplementary.
% VCR dataset contains 290k mutiple-choice QA problems from 110k movie frames. Given an image and a question, four possible answers and four possible rationales for the correct answer are included. There are three tasks available: visual question answering (Q$\rightarrow$A), answer justification (QA$\rightarrow$R) and holistic setting (Q$\rightarrow$AR). The first two tasks can be framed into multiple-choice problems and the holistic setting result can be obtained by combination of first two tasks. We take the output representations of SGEITL and learn a linear classification layer on top to predict a score for each pair. Final prediction is obtained by a softmax function for four possible answers/rationales. 
\vspace{-2mm}
\subsection{Implementation Details} \label{sec4.2}

Details of implementation is attached in the supplementary .

\begin{table}[t]
\begin{center}
\small
\scalebox{0.8}{% 调整表格列间的长度
\begin{tabular}{cccc}
    \toprule
    \multirow{2}{*}{Model} & Number & \multirow{2}{*}{Function $F()$} &  \multirow{2}{*}{Q $\rightarrow$ A} \\
    & of Hops&   &    \\
    \midrule
    w/o HopTrans & - & - & 73.80\\
    \midrule

    \multirow{4}{*}{w/ HopTrans}& 1 & $F(x)=x$ & 73.48 (-0.32)\\
    & 3 & $F(x)=x$ &  73.60 (-0.20)  \\
    & 3 & Gaussian Kernel &  73.89 (+0.09)  \\
    & 3 & Rational Quadratic Kernel & 74.91 (+1.11)  \\
    & 6 & Rational Quadratic Kernel & 74.28 (+0.48)  \\

    \bottomrule
    
\end{tabular}
}
\end{center}
\caption{Ablation on number of hops and different kernel functions in proposed multihop graph Transformer. }
\label{mgt_table}
\vspace{-4mm}
\end{table}

\vspace{-2mm}

\subsection{Ablation Study}

We show the effectiveness of the proposed methods on the validation set of VCR. In Tab. \ref{abalation_table}, we show the experimental results of proposed three components: multihop graph Transformer \textbf{HopTrans}, scene-graph-aware pre-training \textbf{Pretrain-V}  and semantically-relevant scene graphs generated by Text-VSPNet trained by proposed strategy, \textbf{SceneGraph+}. Besides, \textbf{SceneGraph} means relation features generated by Neural Motif \cite{zellers2018scenegraphs} trained on Visual Genome dataset. The baseline of our comparison, the 1st row in table, means that only 36 object features from object detector together with text are inputted into a vanilla Transformer model. 
The baseline model here is a pre-trained VL-BERT~\cite{Su2020VL-BERT:}.

\textbf{Adding Scene Graphs } We first investigate the effect of directly using scene graph and scene graph+ for visual representation. We take the top 18 predicates from both the conventional SGG and our Text-VSPNet according to the predicted confidence score.  Comparing between the 1st, 2nd and 3rd row, we find that solely adding predicate features from both SGGs would already bring around 0.6\% improvement on Q$\rightarrow$A task and  0.9\% on QA$\rightarrow$R task.

\textbf{ Scene Graph-aware Pre-training } Given the merged visual scene graph from Text-VSPNet and the conventional SGG as input, We find it beneficial to replace random masking with triplet-based masking for visual tokens. Compared with the 2nd row, the 3rd experiment with triplet-based masking on visual scene graph boosts the performance of Q$\rightarrow$A and QA$\rightarrow$R by 0.9\% and 0.7\% respectively.

\footnote{It's noted that our Text-VSPNet and the conventional SSG would share the same backbone, object detector thus they would produce two identical set of visual objects but with different predicates. When merging the two types of generated graphs, we simply merge the same type of visual objects and preserve their connected predicates from both graphs.}

\textbf{Multihop Graph Transformer } Furthermore, the proposed multihop graph Transformer is utilized to replace vanilla Transformer model. Through 4th row vs. 3rd row and 6th row vs. 5th row, we demonstrate that multihop graph Transformer can benefit scene graph input with 0.4\% and 0.3\% improvement respectively on Q$\rightarrow$A and QA$\rightarrow$R by incorporating graph structure in Transformer.

To further study the behavior of our framework with multihop graph Transformer, we give a more comprehensive ablation on two parts: multihop and kernel function $F()$. In Tab. \ref{mgt_table}, the baseline (1st row) we use for comparison is the model with ordinary scene graph and triplet-based masking on visual scene graph (Pretrain-V), where the model structure is vanilla Transformer. We find if we only consider one hop neighbor with the identity mapping as $F()$, then the performance drops compared with baseline. If we include multiple hops but still keep identity mapping, as in the 3rd row, the drop of performance gets mitigated a little bit. Then we further substitute identity mapping with learnable Gaussian kernels. This gives us positive result with very marginal improvement. After realizing that the learned Gaussian kernel tends to be very sharp, we replace it with a relatively smoother one - Rational Quadratic kernels and finally obtain satisfactory boost. We also find larger number of hops does not always mean better performance, which might be because that predicates are mostly decided by their close neighbors. Above experiments demonstrate the importance of both multihop information and a suitable kernel. More visualization of learned kernel functions is included in Sec. \ref{sec4.6}.

\textbf{Weakly Supervised Scene Graph Generation}  We study the benefit of proposed weakly-supervised Text-VSPNet generator. Comparing between the 2rd and 3rd row, besides the scene graph generated by a Neural Motif \cite{zellers2018scenegraphs} trained on Visual Genome in fully supervised way, we further incorporate another scene graph from weakly-supervised Text-VSPNet trained on VCR dataset. By providing the predicates that are highly relevant to the text domain of VCR, the performance is further improved by 0.4\% on Q$\rightarrow$A.

% \begin{table}[t]
% \begin{center}
% \renewcommand\tabcolsep{5pt} % 调整表格列间的长度

% \begin{tabular}{ccccc}
%     \toprule
%     \multirow{2}{*}{Model}   & \multirow{2}{*}{SceneGraph} & \multirow{2}{*}{HopTrans} & 
%     \multirow{2}{*}{SceneGraph++} &
%     \multirow{2}{*}{Accuracy} \\
%     &   &  &  \\
%     \midrule
%         Baseline& -  & - & - & 51.63\\
%     \midrule
%     \multirow{2}{*}{SGEITL}& $\surd$  &  & & 52.71 (+1.08)\\
%     & $\surd$ & $\surd$ & & 53.25 (+1.62)  \\

%     \bottomrule
    
% \end{tabular}
% \end{center}
% \caption{Results on the GQA validation set.}
% \vspace{-2mm}
% \label{gqa_table}
% \end{table}

\begin{table}[t]
% \small
\begin{center}
\scalebox{0.7}{
\begin{tabular}{cccccc}
Dataset                                       & Model                   & SceneGraph & HopTrans & SceneGraph+ & Accuracy      \\ \hline
\multicolumn{1}{c|}{\multirow{3}{*}{GQA}}     & Baseline                & -          & -        & -            & 51.63         \\ \cline{2-6} 
\multicolumn{1}{c|}{}                         & \multirow{2}{*}{SGEITL} &  $\surd$          &          &              & 52.71 (+1.08) \\
\multicolumn{1}{c|}{}                         &                         &    $\surd$        &        $\surd$  &              & 53.25 (+1.62) \\ \hline
\multicolumn{1}{c|}{\multirow{4}{*}{SNLI-VE}} & Baseline                & -          & -        & -            & 74.06         \\ \cline{2-6} 
\multicolumn{1}{c|}{}                         & \multirow{3}{*}{SGEITL} &     $\surd$       &          &              & 74.37 (+0.31) \\
\multicolumn{1}{c|}{}                         &                         &      $\surd$      &   $\surd$       &              & 74.83 (+0.77) \\
\multicolumn{1}{c|}{}                         &                         &     $\surd$       &    $\surd$      &   $\surd$           & 75.31 (+1.25) \\ \hline
\end{tabular}
}
\end{center}
\caption{Results on GQA \cite{Hudson_2019_CVPR} and SNLI-VE \cite{xie2018visual} validation set.}
\vspace{-4mm}
\label{gqa_table}
\end{table}

% \vspace{-1mm}
% \subsection{Extensions}
\subsection{Experiments on Other Dataset}

To prove the generalization ability of the proposed framework, we also show some experiments on GQA and SNLI-VE dataset in Tab. \ref{gqa_table}. It is important to note that we focus on validating the generalized advantage of our method across different datasets, so no pre-training or in-depth parameter tuning is conducted, which may make the accuracy score lower than some SOTA methods (triplet-based masking is not included in this experiment). From Tab. \ref{gqa_table}, we can demonstrate the benefits of adding predicate features generated and utilizing graphical structure by multihop graph Transformer. It's noted that the domain of GQA is very close to VisualGenome where \cite{zellers2018scenegraphs} is trained. Both the image and the question all focus on semantically low-level relationships between objects. Thus, SceneGraph+ is not necessary for this task but we do incorporate it in SNLI-VE dataset.

\begin{table}[t]
\begin{center}
\renewcommand\tabcolsep{3pt} % 调整表格列间的长度
\scalebox{0.8}{
\begin{tabular}{ccccc}
    \toprule
    \multirow{2}{*}{model} & \# image-caption &  \multirow{2}{*}{Q$\rightarrow$A} & \multirow{2}{*}{QA$\rightarrow$R} & \multirow{2}{*}{Q$\rightarrow$AR} \\
    & in pre-training& & & \\
    \midrule
    VL-BERT  & 3.3M & 75.5 & 77.9 & 58.9  \\
    UNITER*  & 9.5M & 76.8 & 80.2 & 61.6  \\
    ERNIE-ViL  & 3.8M & 78.9 & 83.7 & 66.4  \\
    
    \midrule
    ERNIE-ViL  & 290k & 74.1 & 76.9 & 56.9  \\
    UNITER  & 290k & 73.4 & 76.0 & 55.8  \\
    \textbf{UNITER+SGEITL} & 290k & \textbf{74.8} & \textbf{76.8} & \textbf{57.4}  \\
    VL-BERT & 290k & 72.9 & 75.3 & 54.9  \\
    \textbf{VL-BERT+SGEITL} & 290k & \textbf{74.9} & \textbf{77.2} & \textbf{57.8}  \\
    \bottomrule
    
\end{tabular}
}
\end{center}
\caption{Comparison with benchmarks (VLBERT \cite{Su2020VL-BERT:}, UNITER \cite{chen2020uniter}, ERNIE-ViL \cite{yu2020ernie-vil}) on VCR validation set. UNITER* denotes that the result is not reported in original paper, but is from official code repository.}

\label{vcr_benchmark}
\vspace{-4mm}
\end{table}
% \vspace{-2mm}

% \footnote{The accuracy readings of VL-BERT and ERNIE-ViL are referenced from the original publication. Since the performance of UNITER on VCR val set is not available, we re-implement its performance here.}
\subsection{Comparison with Benchmarks}
Tab. \ref{vcr_benchmark} shows the comparison between our proposed model and the SOTA methods on the VCR validation set. The baseline here is the original VL-BERT pre-trained on 290k data, referring to the 1st row in Tab. \ref{vcr_benchmark}. And VL-BERT+SGEITL refers to the last row in Tab. \ref{vcr_benchmark}. Compared with baseline, the enhancement of visual scene graph can get a boost of 2.1\% on average in three tasks.  In this work, we focus on proving the advantage of our framework to incorporate visual scene graph on VL models. Our proposed framework is an add-on module that could be applied on top of different Transformer-based VL models. Thus concerning heavy computational cost of generating additional millions of scene graphs, we do not conduct the large-scale pre-training with out-of-domain data as some of the best-performed models on the VCR leaderboard. However, for rigorous fair comparison, we pre-train those top-performed models including UNITER and ERNIE-ViL with the same scale of in-domain data (290k). It turns out that our VL-BERT+SGEITL can still beat powerful SOTA models with the same fair scale of pre-training. These comparisons show that our method is generalizable to other large-scale pre-training scenarios and we leave it for future work.
% that should be generalized well to the larger-scale pre-training scenario.  

\textbf{Submission to VCR Leaderboard:} 
% We also submit our test result. Since the visual feature quality matters a lot for vision-language modelling. 
With this submitted version, we further replace our original object detector with a stronger off-the-shell one \cite{anderson2018bottom}. Then we train our scene graph generator in the same way as before. The test result of VL-BERT+SGEITL with in-domain pre-training and achieve  $76.0 \%$/$78.0 \%$/$59.6 \%$ on Q2A/QA2R/Q2AR of VCR leaderboard under the abbreviation, \textbf{SGEITL}. The performance coincides with the validation results.

% \begin{figure}[t]
% % \vspace{-4mm}
% \begin{center}
% \scriptsize
% \scalebox{0.6}{
%   \includegraphics[width=1\linewidth]{LaTeX/figures/kernel_3181.png}
% }
% \end{center}
% \vspace{-2mm}%Put here to reduce too much white space after your table
%   \caption{The curves correspond to learned kernels from different attention heads. $\alpha$ denotes scale-mixture parameter and $l$ denotes length-scale parameter.}
% \label{fig:kernel}
% \vspace{-4mm}%Put here to reduce too much white space after your table
% \end{figure}
% \vspace{-2mm}
\subsection{Visualization} \label{sec4.6}

\textbf{Learned Kernel Functions} 
Details about visualized learned kernels is attached in the supplementary material.
% In Fig.~\ref{fig:kernel}, learned kernels of different attention heads are shown. We can find that in some heads, the curve is sharp, which means the feature from faraway neighbors is greatly suppressed. On the contrary, some heads show relatively smooth curves, where multiple hops' information are fused. The phenomenon demonstrates that our learnable kernels can adapts to different functionalities depending on the need of attention heads. 

\noindent
\textbf{Generated Semantically-Relevant Scene Graph}
More generated examples are attached in the supplementary material.
% As shown in Fig.~\ref{fig:4}.
% As shown in Fig.~\ref{fig:4}, the generated relations in our improved graph has better semantic relevance compared with traditional scene graph for helping to answer VCR questions. The vocabulary distribution of our improved graph annotation also includes much more "interesting" relations.

\vspace{-3mm}
\section{Conclusion}

We propose a generalized framework to incorporate visual scene graph on top of the current VL models in  pre-training, fine-tuning and model structure levels. Our work is the first to effectively prove that visual scene graph could be useful in pre-training and fine-tuning on highly semantic dataset like VCR. Our multihop graph Transformer helps preserve the graphical structure and our scene graph generator is able to use text annotation with weak supervision to generate domain-relevant scene graph. We present extensive experimental results to prove each proposed component's advantage and compare with the SOTA methods.

\paragraph{Acknowledgement}

This work was supported in part by DARPA MCS program under Cooperative Agreement N66001-19-2-4032. The views expressed are those of the authors and do not reflect the official policy of the Department of Defense or the U.S. Government.
% >>>>>>>>>>>>>>>>>>>>>>>>>>>>>>>>>>>>>>>>>>>>>>>>>>>>>>>>>>>>>>>>>>>>>>>>>>

\bibliography{aaai22}

\clearpage 

\pagebreak

\pagebreak[4]

\newpage

\section{Supplementary Material}
\begin{figure}[htpb]
\begin{center}
\subfigure[Example1]{
\includegraphics[width=8cm]{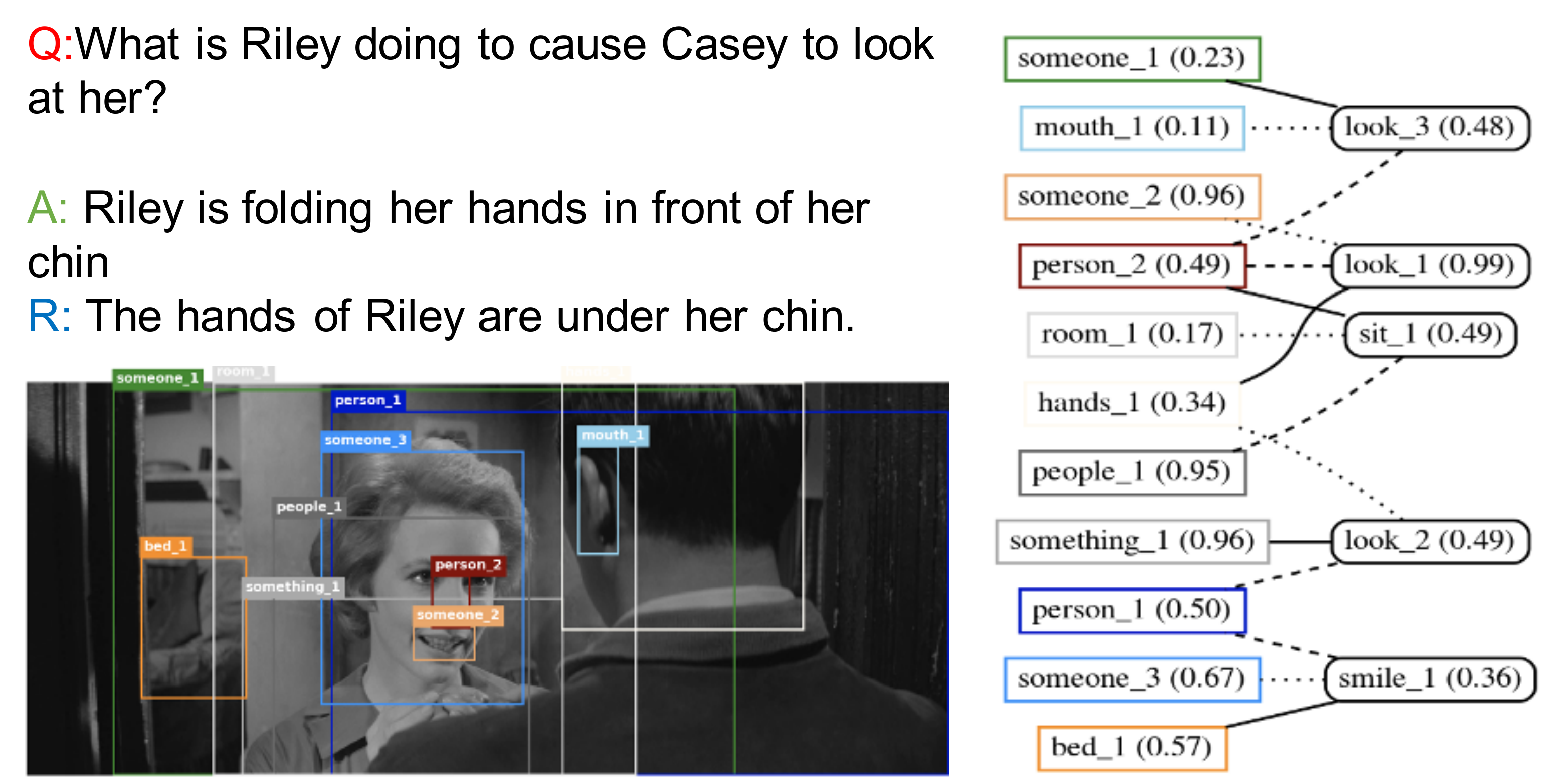}
%\caption{fig1}
}
\quad
\subfigure[Example2]{
\includegraphics[width=8cm]{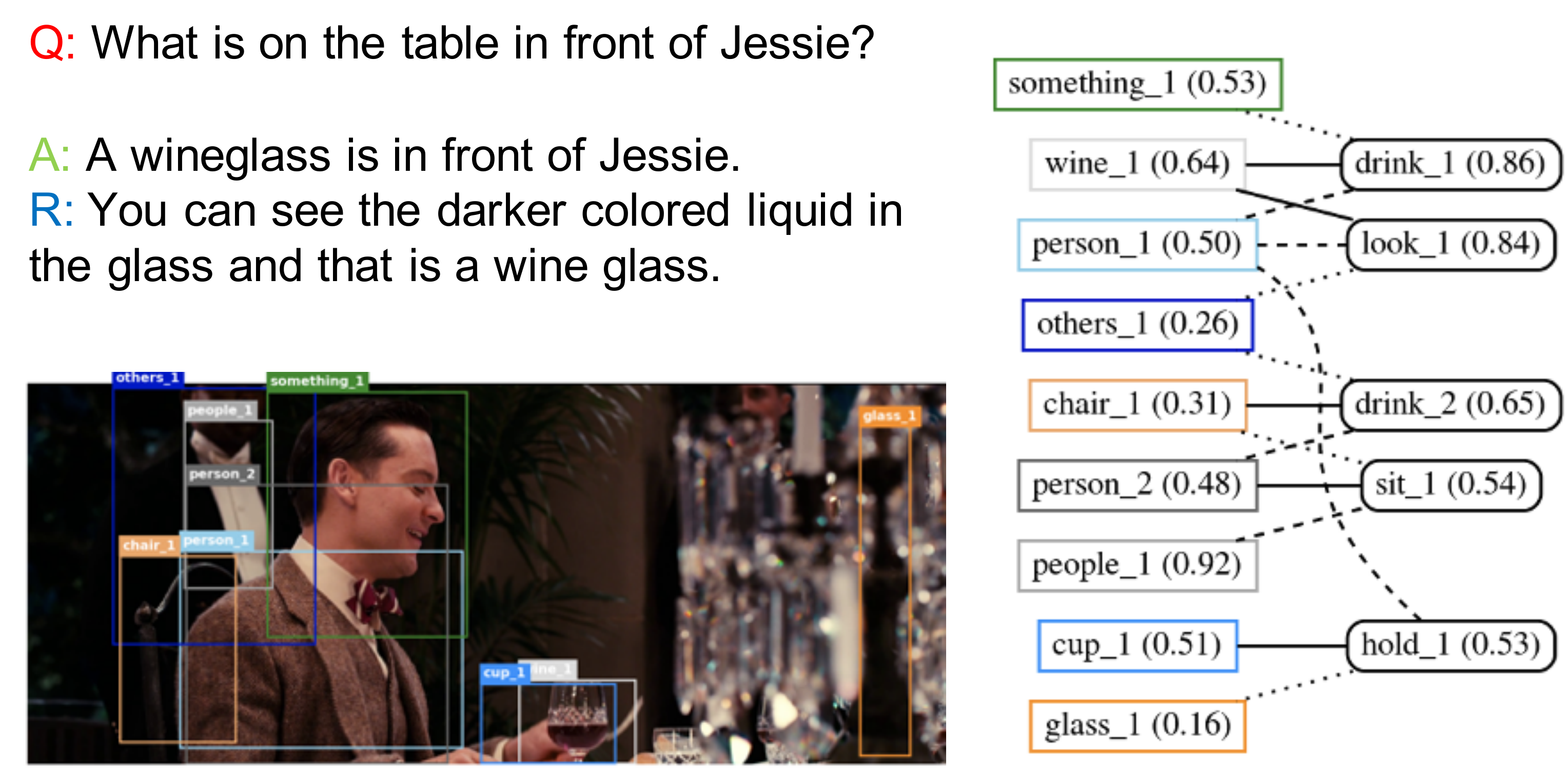}
}
\quad
\subfigure[Example3]{
\includegraphics[width=8cm]{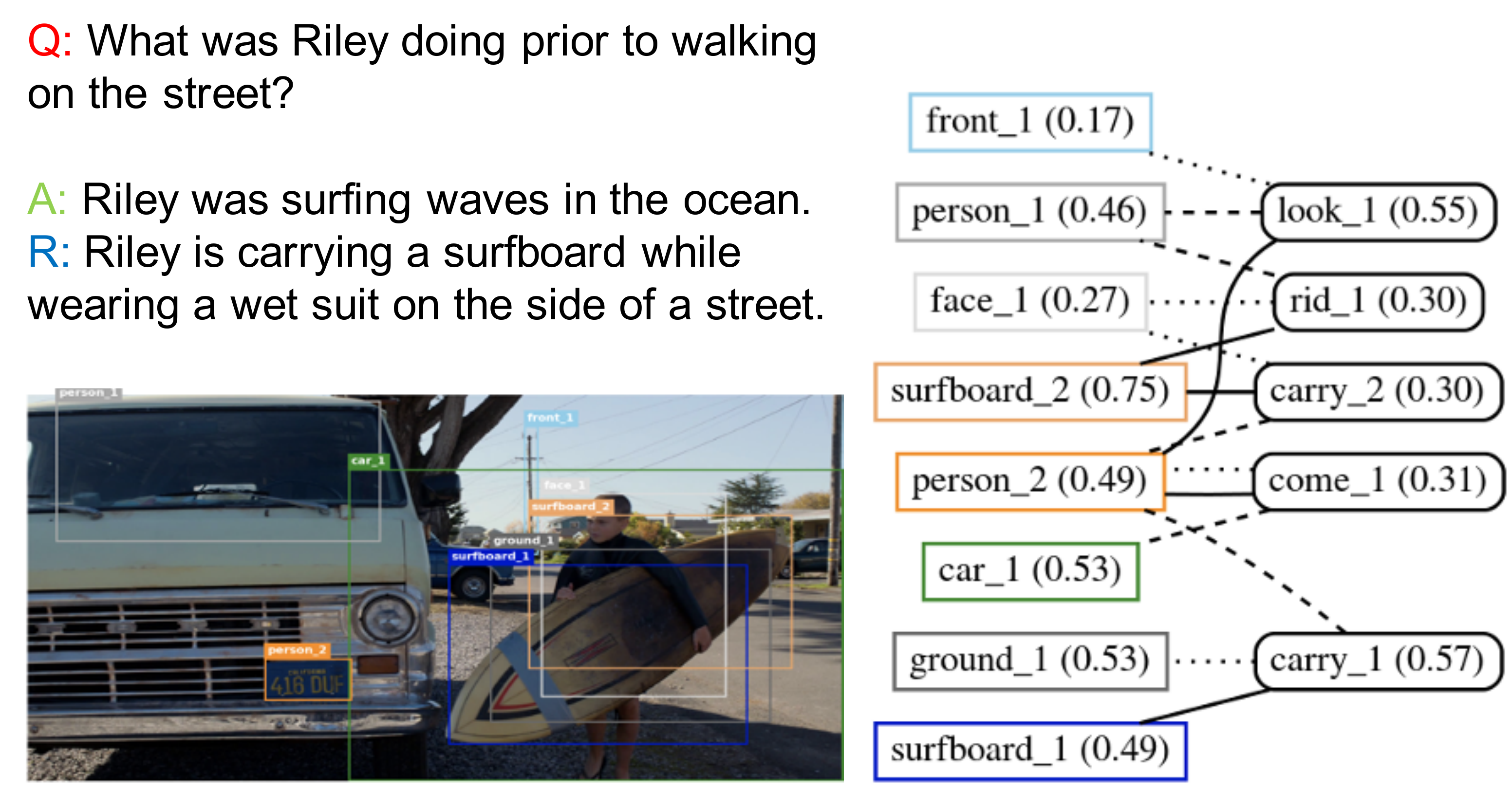}
}
\end{center}
\caption{Here are more examples of semantically relevant scene graphs generated from VSPNet with our proposed training strategy on VCR dataset. Righ-hand Side: Generated scene graphs. Left-hand Side: Questions, golden answers, and golden rationales and images (Displayed images from VCR dataset).}
\label{fig:vspnet}
\end{figure}

\begin{figure}[t]
% \vspace{-4mm}
\begin{center}
\scriptsize
\scalebox{0.6}{
  \includegraphics[width=1\linewidth]{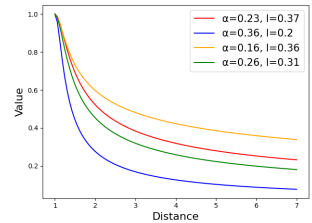}
}
\end{center}
\vspace{-2mm}%Put here to reduce too much white space after your table
  \caption{The curves correspond to learned kernels from different attention heads. $\alpha$ denotes scale-mixture parameter and $l$ denotes length-scale parameter.}
\label{fig:kernel}
\vspace{-4mm}%Put here to reduce too much white space after your table
\end{figure}

% \subsection{Test Result}
% Since the visual feature quality matters a lot for vision-language modelling. We further replace our original object detector with a stronger off-the-shell object detector from \cite{anderson2018bottom}. Then we train our scene graph generator in the same way as before except using above replaced object detector.  We have submitted the new test result to public leaderboard. Since the public leaderboard is maintained by human, we have to wait for several days for them to process and publish. So far, we haven't seen our test accuracy and it is under processing. But it should be public when reviews begins. Here is the link of the public leaderboard: \url{https://visualcommonsense.com/leaderboard/}. \textbf{SGEITL} is the displayed name of our new result, different from previous submission \textbf{GITRL}. We apologize for the potential inconvenience.

\subsection{Dataset}
% Details about dataset is attached in the supplementary material.
VCR dataset contains 290k multiple-choice QA problems from 110k movie frames. The train set contains around 80k images and both the validation and test set contain around 10k images. Based on the original work \cite{zellers2019vcr}, all the three splits are sampled in the same manner. Since the size of both validation and test sets are very similar and based on our observation of past high-performing models' results on both sets, models' performance on validation set can also reflect its performance on test set. Given an image and a question, four possible answers and four possible rationales for the correct answer are included. There are three tasks available: visual question answering (Q$\rightarrow$A), answer justification (QA$\rightarrow$R) and holistic setting (Q$\rightarrow$AR). The first two tasks can be framed into multiple-choice problems and the holistic setting result can be obtained by combining the results of first two tasks. We take the output representations of SGEITL and learn a linear classification layer on top to predict a score for each pair. Final prediction is obtained by a softmax function for four possible answers/rationales. 

\subsection{Implementation Details} 

% Details about implementation is attached in the supplementary material.
% \textbf{Visual Features} Before pre-training, we initialize the parameters of Transformers with $\text{BERT}_{\text{large}}$ pre-trained weight and leave others randomly initialized. Visual object feature is produced by a Faster R-CNN, initialized from parameters pre-trained on Visual Genome (Krishna et al., 2017) for object detection. For the visual predicate features, we take the features before the predicate classification layer in scene graph generators, where their object detectors share the same parameter as the Faster R-CNN we used to extract object feature. 

\textbf{Pre-training} In recent Vision-Language models, outstanding performance improvement in multimodal tasks such as VQA \cite{VQA}, GQA \cite{Hudson_2019_CVPR} and Image Retrieval \cite{Johnson_2015_CVPR} heavily relies on largely pre-training on large-scale image and text dataset. In UNITER \cite{chen2020uniter}, the authors find through experiments that with pre-training solely on in-domain data of VCR, UNITER can also achieve decent performance boost. Incorporating visual scene graph in large pre-training with out of domain data as in \cite{yu2020ernie-vil, chen2020uniter} requires heavily computational cost for generating visual scene graph for millions of image-text data on the first place. This extremely high cost is unrealistic for academic research. Since the focus of this paper is to prove our generalized framework's advantage on incorporating visual scene graph to improve current VL models in pre-training, fine-tuning and model structure, thus such heavy computation is less relevant. In our experiments, we \textbf{only conduct pre-training with in-domain data of VCR with 290k image-caption pairs}.

\textbf{Fine-tuning} For fine-tuning, we first feed the image into the conventional SGG \cite{zellers2018scenegraphs} and our Weakly-supervised Text-VSPNet to obtain an initial combined scene graph of each image. Consequently, the processed scene graph and text information are inputted into our pre-trained SGEITL and the possibility of each choice being correct is obtained.  We  train  our  model  for  20 epochs with SGD optimizer. Initial learning rate is 0.05 and will decay  by  0.1  at  14th  and  18th  epoch. BERT-Large  setting(1024 dim) is used in SGEITL and other compared models.

\begin{figure}[t]
\centering
\includegraphics[width=8cm]{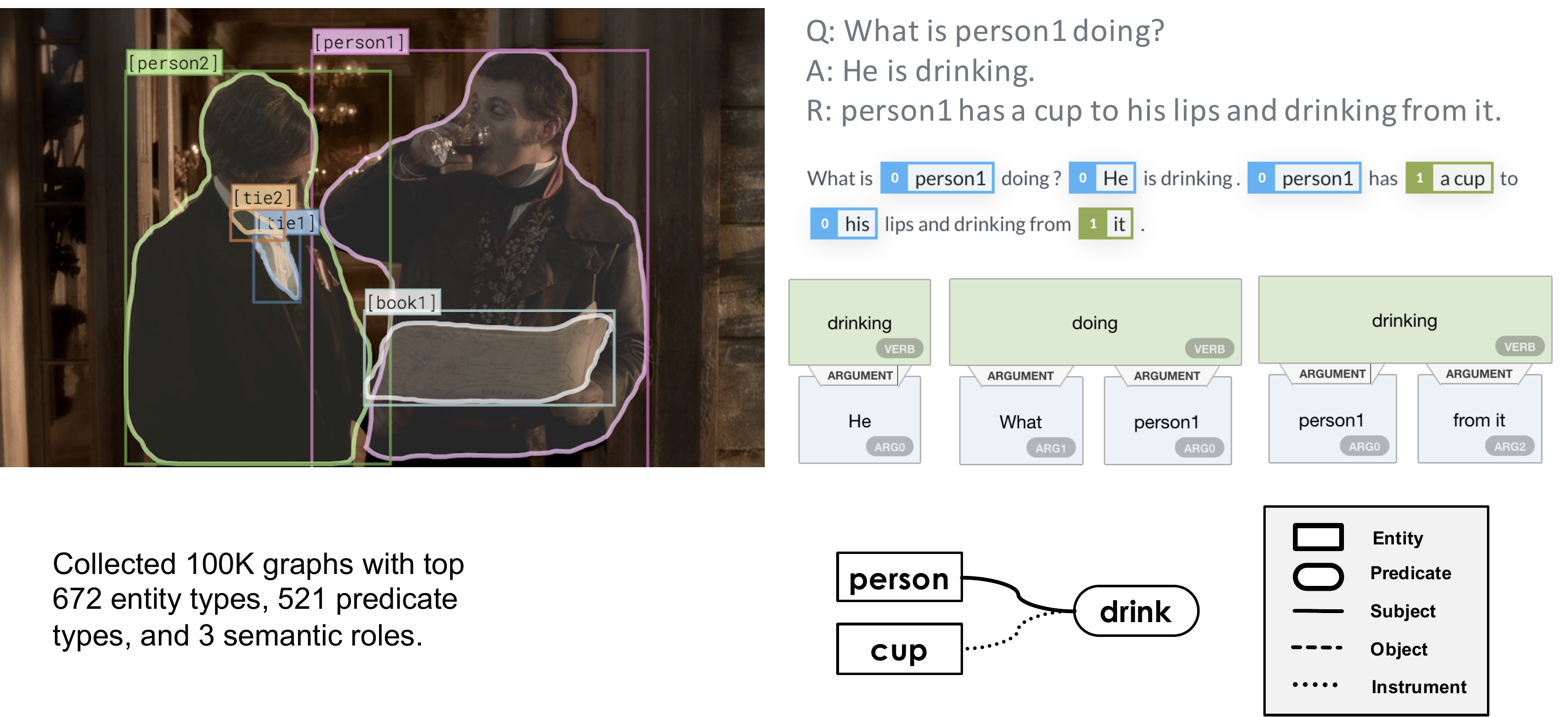}
%\caption{fig1}
\caption{An example of extracted graph on VCR data. We use a semantic parser to extract verbs, nouns, and semantic roles from each question, answer, and rationale (displayed image from VCR dataset).}
\label{fig:srp}
\end{figure}

\textbf{VSPNet Training }  
We extract VCR Scene Graph annotation, which consists of entities and their relations. We first concatenate sentences of question with correct answer and rationale from the VCR dataset, and use end-to-end neural co-reference solution \cite{lee2017end} to match expressions in concatenated sentences that refer to the same entities and BERT-based Semantic Role Labeler (SRL) to extract the SRL graphs, which assigns semantic roles to phrases \cite{shi2019simple}, as illustrated in Fig. \ref{fig:srp}. We filter SRL graphs by keeping only the top 4 most frequent semantic roles, which are action (V), agent (ARG0), patient (ARG1), as well as instrument, benefactive, or attribute (ARG2). Among the top-20 most frequent lemmatized verb expressions, we remove graphs with verbs that are too abstract to be visually distinguished, which means the verbs are irrelevant to the image, incorrectly extracted, or are auxiliary verbs. After filtering to keep the predicates and entities with frequency of at least 100 occurrences, 99\% of the images and 90\% of QAR annotations in the training set are preserved, and 657 different entity types and 521 different predicate types are extracted. More generated scene graphs are shown in Fig. \ref{fig:vspnet}

\subsection{Visualization}

\textbf{Learned Kernel Functions} 
% Details about visualized learned kernels is attached in the supplementary material.
In Fig.~\ref{fig:kernel}, learned kernels of different attention heads are shown. We can find that in some heads, the curve is sharp, which means the feature from faraway neighbors is greatly suppressed. On the contrary, some heads show relatively smooth curves, where multiple hops' information is fused. The phenomenon demonstrates that our learnable kernels can adapt to different functionalities depending on the need of attention heads. 

\noindent
\textbf{Generated Semantically-Relevant Scene Graph}
% More generated examples are attached in the supplementary material.
As shown in Fig.~\ref{fig:vspnet}, the generated relations in our improved graph have better semantic relevance compared with traditional scene graph for helping to answer VCR questions. The vocabulary distribution of our improved graph annotation also includes much more "meaningful" relations.

% >>>>>>>>>>>>>>>>>>>>>>>>>>>>>>>>>>>>>>>>>>>>>>>>>>>>>>>>>>>>>>>>>>>>>>>>>>

% \bibliography{aaai22}

\end{document}